\documentclass[conference]{IEEEtran}
\IEEEoverridecommandlockouts
\usepackage{cite}
\usepackage{amsmath,amssymb,amsfonts}
\usepackage{algorithmic}
\usepackage{graphicx}
\usepackage{textcomp}
\usepackage{xcolor}
\usepackage{diagbox}
\usepackage{subcaption}
\usepackage{comment}
\usepackage{url}
\usepackage{tikz}
\usepackage[absolute,overlay]{textpos}
\usepackage{blindtext}
\usepackage{relsize}
\usepackage{hyperref}

\tikzset{fontscale/.style={font=\relsize{#1}}}

\def\BibTeX{{\rm B\kern-.05em{\sc i\kern-.025em b}\kern-.08em
    T\kern-.1667em\lower.7ex\hbox{E}\kern-.125emX}}
\begin{document}

\title{Designing DNNs for a trade-off between robustness and processing performance in embedded devices*\\
\thanks{*This work was partially supported by the University of the Basque Country (UPV-EHU) under grant GIU21/007, by the Basque Government under grants PRE\_2023\_2\_0148 and KK-2023/00090 and by the Spanish Ministry of Science and Innovation under grant PID2020-115375RB-I00. \\}
}

\author{\IEEEauthorblockN{1\textsuperscript{st} Jon Gutiérrez-Zaballa}
\IEEEauthorblockA{\textit{Dept. of Electronics Technology} \\
\textit{University of the Basque Country}\\
Bilbao, Spain \\
j.gutierrez@ehu.eus}
\and
\IEEEauthorblockN{2\textsuperscript{nd} Koldo Basterretxea}
\IEEEauthorblockA{\textit{Dept. of Electronics Technology} \\
\textit{University of the Basque Country}\\
Bilbao, Spain \\
koldo.basterretxea@ehu.eus}
\and
\IEEEauthorblockN{3\textsuperscript{rd} Javier Echanobe}
\IEEEauthorblockA{\textit{Dept. of Electricity and Electronics} \\
\textit{University of the Basque Country}\\
Leioa, Spain \\
franciscojavier.echanove@ehu.eus}
}

\maketitle

\begin{textblock*}{21cm}(1.5cm,26cm)
\begin{tikzpicture}
    \draw (0,0) rectangle (18,0.5); 
    \end{tikzpicture}
\end{textblock*} 

\begin{textblock*}{21cm}(0cm,26cm)
\begin{tikzpicture}
    \node (center) {c};
    \path (center)+(10.5,4) node [fontscale=-1] (name) {\copyright 2024 IEEE. Final published version of the article can be found at \href{https://ieeexplore.ieee.org/document/10769119}{10.1109/DCIS62603.2024.10769119}.};
    \end{tikzpicture}
\end{textblock*} 

\begin{abstract}
Machine learning-based embedded systems employed in safety-critical applications such as aerospace and autonomous driving need to be robust against perturbations produced by soft errors.
Soft errors are an increasing concern in modern digital processors since smaller transistor geometries and lower voltages give electronic devices a higher sensitivity to background radiation.
The resilience of deep neural network (DNN) models to perturbations in their parameters is determined, to a large extent, by the structure of the model itself, and also by the selected numerical representation and used arithmetic precision.
When compression techniques such as model pruning and model quantization are applied to reduce memory footprint and computational complexity for deployment, both model structure and numerical representation are modified and thus, soft error robustness also changes.
In this sense, although the choice of activation functions (AFs) in DNN models is frequently ignored, it conditions not only their accuracy and trainability, but also compressibility rates and numerical robustness.
This paper investigates the suitability of using bounded AFs to improve model robustness against DNN parameter perturbations, assessing at the same time the impact of this choice on deployment in terms of model accuracy, compressibility, and computational burden.
In particular, we analyze encoder-decoder fully convolutional models aimed at performing semantic segmentation tasks on hyperspectral images for scene understanding in autonomous driving.
Deployment characterization is performed experimentally on an AMD-Xilinx's KV260 SoM.
\end{abstract}


\section{Introduction}\label{sec:IntroRelatedWork}
Deploying AI accelerators on the edge help to reduce issues related to security, reliability, and high response latencies by avoiding the communication of data from/to external, more powerful computing platforms.
Moreover, by providing full execution autonomy to complex AI algorithms such as deep neural networks (DNNs), AI can be applied to tasks with strict communication, reliability, and real-time response requirements. 
Computer vision-based intelligent systems for autonomous driving systems (ADSs) and aerospace applications are two relevant examples with high economic impact.
However, the success of DNNs in many applications very often comes at the cost of designing highly complex models with millions of parameters that require tens of giga floating-point operations (GFLOPS) per inference.
Deploying such models on resource-constrained embedded AI processors inevitably requires applying certain compression techniques and addressing a trade-off between accuracy, hardware occupation, and inference speed.

In addition to the above, a major concern in the deployment of AI-based autonomous systems is the requirement of meeting strict safety and reliability standards, especially in the aerospace (ARP-4754) and automotive (ISO 26262) industries,
One of the main factors that jeopardize the reliability of such systems is the exposure of electronic components to background radiation.
This is particularly true for memory devices that rely on storing small amounts of charge and that also occupy large proportions of total silicon area.
In the simplest case, the logical value stored in a cell may be perturbed, resulting in a single event upset (SEU) or a single bit upset (SBU) if just a single bit is altered.
Field programmable gate arrays (FPGAs), which are increasingly being used in these fields, are particularly sensitive to SBUs, since both the configuration memory (LUTs) and the sequential elements of the deployed circuit (flip-flops or Block RAMs) can be affected.
Without sufficient protection against soft errors, the mean time between failures could be seconds.

Most published papers on the field have focused on image classification tasks, with little attention to semantic segmentation models.
The most widely used approach to assess DNN robustness is through simulated software fault injection (FI) campaigns, while only a few authors report a theoretical analysis based on vulnerability models \cite{systematicReview}.
One of the most exhaustive analyses of the vulnerability of 32-bit floating-point convolutional neural networks for image classification is described in \cite{hong2019terminal}.
The key findings are that most of errors come from drastic spikes in parameter values, with positive ones being more threatening, and that dropout and batch normalization layers are ineffective in preventing error propagation.
It is also worth mentioning the work presented in \cite{narayanan2021fault}, in which two software tools for exhaustive FI in both \textit{TensorFlow1} and \textit{TensorFlow2} \cite{tensorfi2} frameworks are explained.

Regarding methods to harden DNNs against soft errors, the main approaches include redundancy, parameter modification, changes in the training/inference strategy, and reshaping of activation functions (AFs).
However, most of them require additional computing resources, making its applicability to embedded systems difficult.
Particularly interesting is the work in \cite{jang2021mate}, where the authors observe that the least significant bits of the mantissa of the parameter values are weakly linked to accuracy.
Accordingly, the authors present MATE, an error correction tool with no memory overhead based on the substitution of those bits by error correction codes for the weights.
A different methodology is presented in \cite{taheri2024exploration}, where the maximum values of the AFs are evaluated and then replaced with either lower or upper-bound values to reduce the propagation of errors through layers.

In this work we study the suitability of using bounded AFs as a means to improve the robustness of image segmentation DNNs and how this choice can impact the performance and implementability of these models.
With this aim, we analyze the process of deploying an encoder-decoder DNN on an FPGA for a real-world task: the semantic segmentation of hyperspectral images in the context of autonomous driving.

Section \ref{sec:modelDevelopment} describes how the DNN under study has been developed in terms of selection of the AFs, training, pruning, and quantization for deployment on edge devices.
In Section \ref{sec:robustness}, we analyze the effects of applied compression techniques on model robustness according to the choice of the AFs by means of an statistically significant FI campaign.
In Section \ref{sec:characterization}, we give details about the deployment of the models on an FPGA and report comparative performance figures.
Finally, the concluding remarks are presented in Section \ref{sec:conclusions}.

\section{Models' development}\label{sec:modelDevelopment}
The reference segmentation model in this study is a U-Net, an encoder-decoder fully convolutional network for image segmentation tasks.
This DNN has been adapted to use hyperspectral images (HSI) from HSI-Drive v2.0 \cite{gutierrez2023hsi}, a dataset intended for developing ADS systems using HSI.
The most recent version, described in \cite{GUTIERREZZABALLA2024103242}, is based on a 5-level encoder-decoder architecture containing two sequences of 3x3 $conv2D$ layers (initially with 32 filters) followed by Batch Normalization and Rectified Linear Unit (ReLU) AFs at each level.
Additionally, it includes one 2x2 Max-pooling 2D layer per encoder level and one 2x2 $conv2D_{tr}$ layer per decoder level.
The resulting model features 31.14 million parameters and requires 34.60 GFLOPS per inference to execute.

Since the lack of bounds in most widely used AFs for DNN implementation (e.g. ReLU) facilitates the propagation of soft errors, in this article we also explore the use of squashing AFs, such as Sigmoid and Hard Sigmoid, being the latter a more basic, computationally efficient version of the former \cite{APICELLA202114}.
The objective is to analyze their potential benefits on model robustness, while also assessing their impact on performance and implementability.
The DNNs have been designed with \textit{TensorFlow2} and trained on a Dell Precision 7920 Workstation equipped with an NVIDIA RTX 3090 GPU.

\subsection{Reference model's performance}
First, the three AFs are evaluated based on their performance in training and testing using the reference noncompressed model.
The Sigmoid-based and Hard Sigmoid-based models required 1000 epochs, while the ReLU-based DNN, which allows for faster convergence, was trained for only 200 epochs.
Table \ref{tab:metricsQuantizedUnprunedPrunedModel} shows the best Intersection Over Union (IoU) results of the 32-bit floating-point DNNs on the test set according to the used AF.

\begin{table}[t]
\caption{IoU of 32-bit floating-point DNNs on the test set.}
\label{tab:metricsQuantizedUnprunedPrunedModel}
\centering
\begin{tabular}{c|c|c|c|}
\hline
\multicolumn{1}{|c|}{\diagbox[]{\textbf{Class}}{\textbf{AF}}} & \multicolumn{1}{c|}{\textbf{ReLU}} & \multicolumn{1}{c|}{\textbf{Sigmoid}} & \multicolumn{1}{c|}{\textbf{Hard Sigmoid}} \\ \hline
\multicolumn{1}{|c|}{\textbf{Road}}       & \multicolumn{1}{c|}{97.84} & \multicolumn{1}{c|}{96.94} & \multicolumn{1}{c|}{96.32} \\ \hline
\multicolumn{1}{|c|}{\textbf{Marks}}      & \multicolumn{1}{c|}{87.99} & \multicolumn{1}{c|}{82.88} & \multicolumn{1}{c|}{83.43} \\ \hline
\multicolumn{1}{|c|}{\textbf{Vegetation}} & \multicolumn{1}{c|}{94.23} & \multicolumn{1}{c|}{92.53} & \multicolumn{1}{c|}{92.39} \\ \hline
\multicolumn{1}{|c|}{\textbf{Sky}}        & \multicolumn{1}{c|}{92.83} & \multicolumn{1}{c|}{89.47} & \multicolumn{1}{c|}{85.17} \\ \hline
\multicolumn{1}{|c|}{\textbf{Others}}     & \multicolumn{1}{c|}{78.12} & \multicolumn{1}{c|}{76.38} & \multicolumn{1}{c|}{69.07} \\ \hline
\multicolumn{1}{|c|}{\textbf{Global}}     & \multicolumn{1}{c|}{94.71} & \multicolumn{1}{c|}{93.32} & \multicolumn{1}{c|}{92.02} \\ \hline
\multicolumn{1}{|c|}{\textbf{Weighted}}   & \multicolumn{1}{c|}{88.54} & \multicolumn{1}{c|}{84.75} & \multicolumn{1}{c|}{82.69} \\ \hline
\end{tabular}
\end{table}

The best metrics are obtained for the ReLU model, although in all three cases, the Global IoU (GIoU) is above 92\%.
The Weighted IoU (WIoU), which is calculated by weighting each class by the inverse of its frequency in the dataset, is above 82\%.
These results are considered satisfactory given the highly unbalanced nature of the dataset \cite{gutierrez2023hsi}.

\subsection{Model pruning}\label{sec:pruning}
For this model to be implemented on an edge device, it is necessary to apply certain compression techniques such as pruning and quantization.
Channel pruning aims to reduce both the number of parameters and the number of FLOPS by removing channels that have minimal impact on the output.
This is achieved through a model sensitivity analysis which consists of gradually pruning (in increments of 0.1 in our case) each of the parameters, while the rest of the model remains frozen, to estimate which layers are the least essential ones.
After choosing an overall pruning ratio in terms of FLOPS, each of the layers is pruned accordingly, and then the DNN is fine-tuned for a certain number of epochs (60 for ReLU and 200 for the other two AFs) to recover any lost accuracy.

This process is repeated twice for each model in what is known as iterative pruning, thus requiring the whole process to be repeated on the pruned model after the first iteration.
To consider pruning as valid, a maximum degradation of 1.5 points in both GIoU and WIoU has been accepted, ensuring both remain above 90\% and 80\%, respectively.

\begin{figure}[t]
\centerline{\includegraphics[height = 5cm]{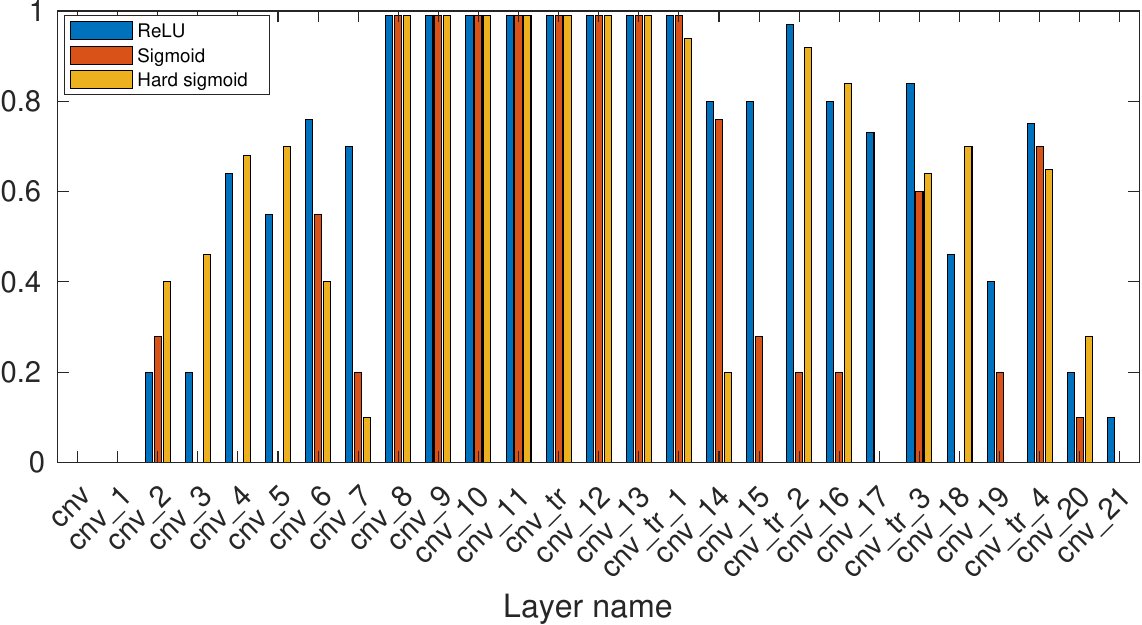}}
\caption{Overall pruning ratio in each of the layers in the models with ReLU (0.5 and 0.5), Sigmoid (0.4 and 0.2) or Hard Sigmoid (0.5 and 0.4) AFs.}
\label{fig:overallPruningRatio}
\end{figure}

The overall pruning ratios have been: 0.75 (0.5 and 0.5) for the ReLU-based model, 0.52 (0.4 and 0.2) for the Sigmoid-based model, and 0.7 (0.5 and 0.4) for the Hard Sigmoid-based model.
Fig. \ref{fig:overallPruningRatio} illustrates the pruning ratio applied to each layer, where it can be noted that, even though the overall pruning ratio of the ReLU-based DNN and the Hard Sigmoid-based model are very similar, the individual pruning ratio greatly varies from layer to layer, especially in the initial layers of the encoder and the final layers of the decoder.
It can also be seen that, from layer $cnv\_6$ to layer $cnv\_15$, which are the ones placed around the base, the Sigmoid-based DNN is more pruned than the Hard Sigmoid-based DNN.

As the layers next to the base of the architecture are the ones that contain most of the parameters, even though the overall pruning ratio in terms of FLOPS is smaller for the Sigmoid-based DNN, it results in a model containing fewer parameters than the Hard Sigmoid-based DNN.
Table \ref{tab:modelSize} shows the complexity and size of the models, which have been significantly reduced after the pruning process.

\subsection{Quantization}\label{sec:quantization}
Quantization is another compression technique aimed at reducing the size of the parameters to store and accelerating model inference when deployed on customized hardware.
In this article, we apply a homogeneous post-training quantization scheme where all parameters and activations are converted to 8-bit integers.
For further details on the quantization process, the reader is referred to \cite{10382745}.
It is worth noting that after quantization, the Sigmoid-based model has experienced a noticeable degradation.
To recover the lost accuracy, a process called fast finetuning has been carried out.

\begin{table}[b]
\caption{DNN complexity and size reduction after applying channel-pruning and quantization.}
\label{tab:modelSize}
\centering
\begin{tabular}{|c|c|c|c|c|}
\hline
\textbf{\diagbox[]{Metric}{AF}} & \textbf{Original}  & \textbf{ReLU} & \textbf{Sigmoid}        & \textbf{Hard Sigmoid} \\ \hline
\textbf{Params (M)}               &   31.13            &   0.32        &   1.38                  &         1.47          \\ \hline
\textbf{OPS (G)}                  &   34.59            &   8.41        &  16.46                  &         10.35         \\ \hline
\textbf{Memory (MB)}              &   249.04           &   2.49        &  11.04                  &         11.76         \\ \hline
\end{tabular}
\end{table}

\section{Analysis of robustness against SBUs}\label{sec:robustness}
To test the models' robustness against SBUs, an extensive FI campaign was conducted on the aforementioned models, which involved injecting single bit-flips into the parameters of the DNNs.
Perturbations were inserted using our modified version of the original \textit{TensorFI2} framework \cite{tensorfi2}, and the code is shared at \textit{\url{https://github.com/jonGuti13/TensorFI2}}.
To ensure the statistical significance, which is very important as stated in \cite{ruospo2023assessing}, of the performed FI campaign 1550 different soft errors per parameter set (a 2.5\% error margin, a 95\% confidence level, and a 50\% failure probability to maximize sample size \cite{statisticalFaultInjection}) have been injected.
To evaluate the impact of the FI campaign, it is first necessary to define what an inference error is.
A bit-flip is considered to have caused an error if the predicted class at any pixel in the test images changes with regard to the prediction made by the unperturbed original model, i.e., critical errors.
The error rate is therefore presented as a percentage between 0 and 100, with 100 representing the situation where a bit-flip has changed the predicted class of every single pixel.

\subsection{Original noncompressed models}
\begin{figure}[b]
\centerline{\includegraphics[height = 6.75cm]{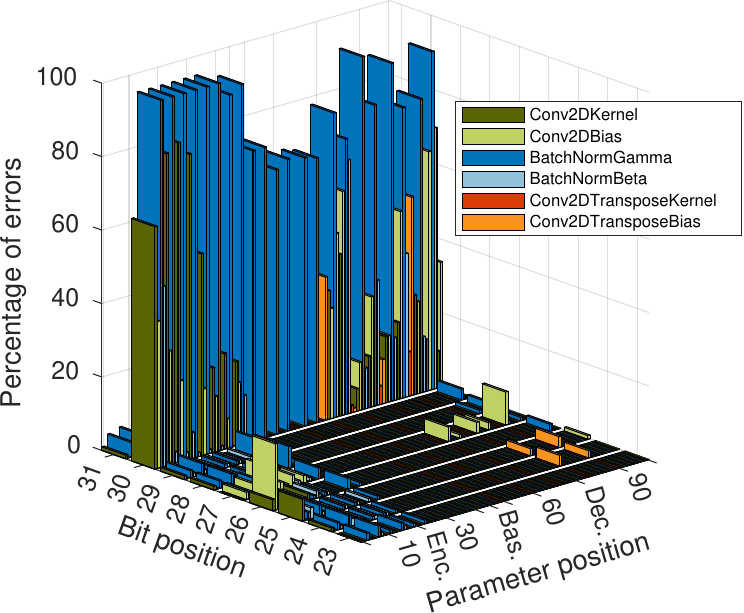}}
\caption{Bit-flip error rate for the original ReLU-based DNN.}
\label{fig:bitFlipErrorNotPruned}
\end{figure}

Fig. \ref{fig:bitFlipErrorNotPruned} displays the error rate for each parameter set and flipped bit for the ReLU-based 32-bit floating-point DNN (for an in-depth analysis of this model the reader is referred to \cite{GUTIERREZZABALLA2024103242}).
Only sign and exponent bits are shown because perturbations in mantissa bits generate a negligible amount of errors.
Even though Fig. \ref{fig:bitFlipErrorNotPruned} varies for each of the nonquantized DNNs under study (nonpruned/pruned and ReLU/Sigmoid/HardSigmoid), they also share a common pattern.

Firstly, errors primarily occur as a consequence of the increment of the perturbed parameters' values.
Secondly, the most sensitive bit is the MSB bit of the exponent as it is originally a '0' in most of the DNNs parameters (the ones in $0 \leq |x| < 2$ range).
In fact, if the original parameter value is $1$ or is in $1 < |x| < 2$ range, a bit-flip converts the parameter into a $\pm \infty$ or a $NaN$, respectively.
Thirdly, the most sensitive parameter is the gamma set of Batch Normalization layers as it usually has a value near $1$.
Finally, leaving aside the MSB, the most sensitive zones are the initial layers of the encoder and the final layers of the decoder, which are both directly connected to the output as a consequence of the skip-connections typical of encoder-decoder architectures.

\begin{figure}[b]
\centerline{\includegraphics[height = 6cm]{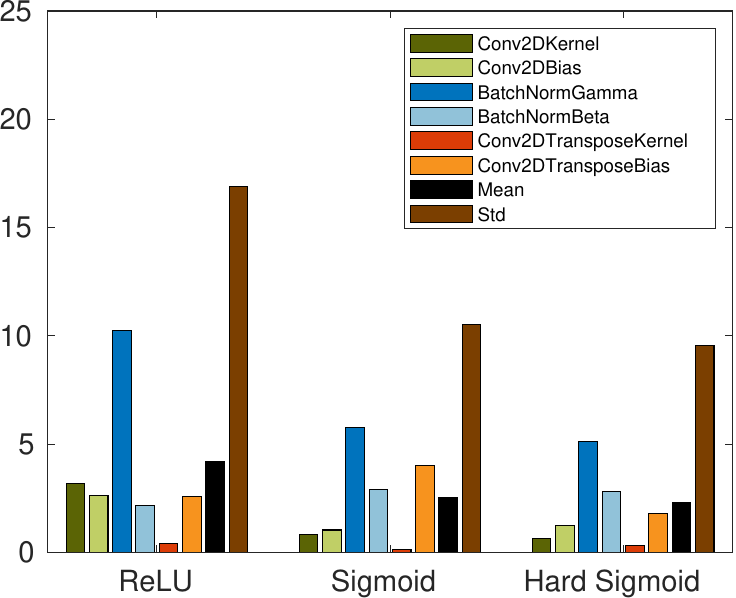}}
\caption{Mean bit-flip error rate in the original model.}
\label{fig:comparativaErrorRateGeneralNotPruned}
\end{figure}

Fig. \ref{fig:comparativaErrorRateGeneralNotPruned} displays a comparison of measured mean error rates of each parameter set for the three AFs.
The error rates are smaller for the DNNs with squashed AFs, especially in the kernel/bias parameters of the $Conv2D$ layers, benefiting from the fact that faulty parameter values cannot grow indefinitely, and neither can the error.
The Hard-Sigmoid-based model exhibits the highest level of robustness.

\begin{figure}[t]
\centerline{\includegraphics[height = 7cm]{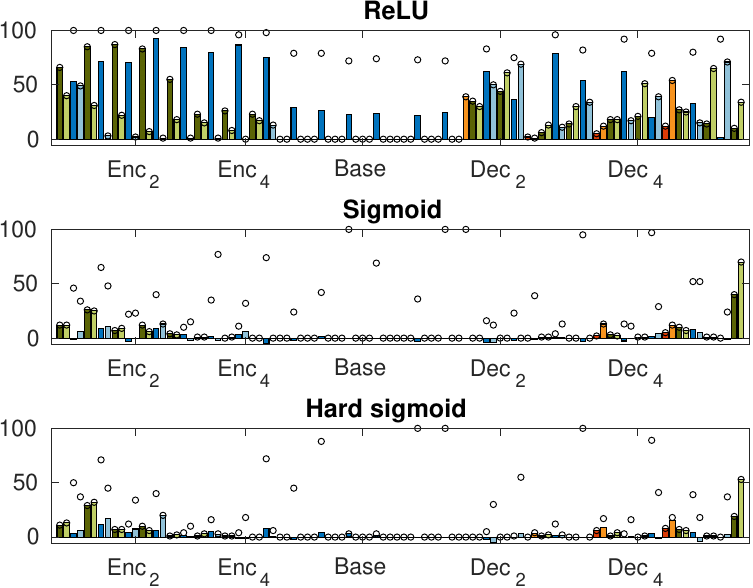}}
\caption{Difference between MSB bit-flip error rate and ratio of parameter set values in $1 < |x| < 2$. Unfilled dots indicate MSB bit-flip error rate. Original model.}
\label{fig:comparativaErrorRateMSBNotPruned}
\end{figure}

Another aspect that differs as a consequence of the AF used is the range of parameter values.
As explained, a range considered to be potentially dangerous is $1 < |x| < 2$.
In Fig. \ref{fig:comparativaErrorRateMSBNotPruned}, the difference between the MSB bit-flip error rate and the ratio of parameter set values in $1 < |x| < 2$ is displayed.
Minor negative peaks are a consequence of the statistical FI campaign where not all the parameters per set are perturbed.

Fig. \ref{fig:comparativaErrorRateMSBNotPruned} shows how the MSB bit-flip errors in the central area of the U-Net, which is composed of the last layers of the encoder, the base, and the first layers of the decoder, mainly occur because of the MSB bit-flip of the parameters which are in the $1 < |x| < 2$ interval.
At the same time, we can also see the benefits of using bounded AFs as the MSB bit-flip error rate is smaller than in ReLU based DNN.

\subsection{Pruned models}
\begin{figure}[b]
\centerline{\includegraphics[height = 6cm]{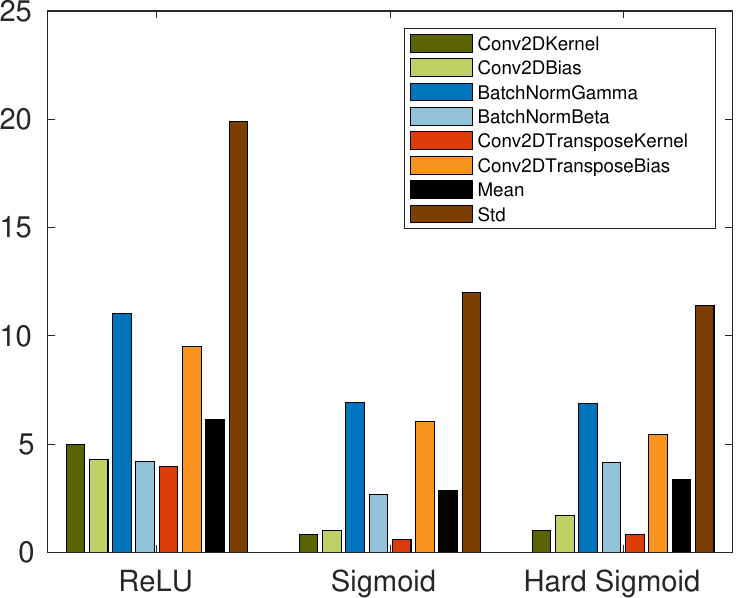}}
\caption{Mean bit-flip error rate in the pruned model.}
\label{fig:comparativaErrorRateGeneralPruned}
\end{figure}

In this subsection it is analyzed to what extent the vulnerability of the DNNs against SBUs changes as a consequence of the pruning process described in Section \ref{sec:pruning}.
From the comparison between Fig. \ref{fig:comparativaErrorRateGeneralNotPruned} and Fig. \ref{fig:comparativaErrorRateGeneralPruned} it can be concluded that the error rate has augmented after the DNNs have been pruned.
However, the three models have not experienced an equal loss of robustness.
The more over-parameterized a model is, the greater the likelihood that a bit-flip will occur on a less critical parameter, thus not impacting the predicted classes.
Indeed, the two most pruned models (ReLU and Hard Sigmoid) have shown the most degradation, while the Sigmoid-based DNN now exhibits the highest robustness, surpassing the Hard-Sigmoid in terms of mean bit-flip error rate (as indicated by the black bar in Fig. \ref{fig:comparativaErrorRateGeneralPruned}).
Nevertheless, albeit marginally, the Hard-Sigmoid demonstrates a lower standard deviation, suggesting less variability in the bit-flip error rate across different positions.

From this analysis, it may seem that pruning is only recommended for reducing the complexity of DNNs and accelerating their inference, and that it has detrimental effects on their robustness.
However, in most implementations, the probability of a SEU occurring in smaller models is also lower due to decreased device occupation, so to accurately assess the influence of pruning on vulnerability to SEUs, factors such as circuit design and chosen target device must be considered.

Pruning also increases the proportion of values in the range $1 < |x| < 2$ by removing channels with parameters that are usually close to 0 and considered irrelevant.

\begin{figure}[t]
\centerline{\includegraphics[height = 7cm]{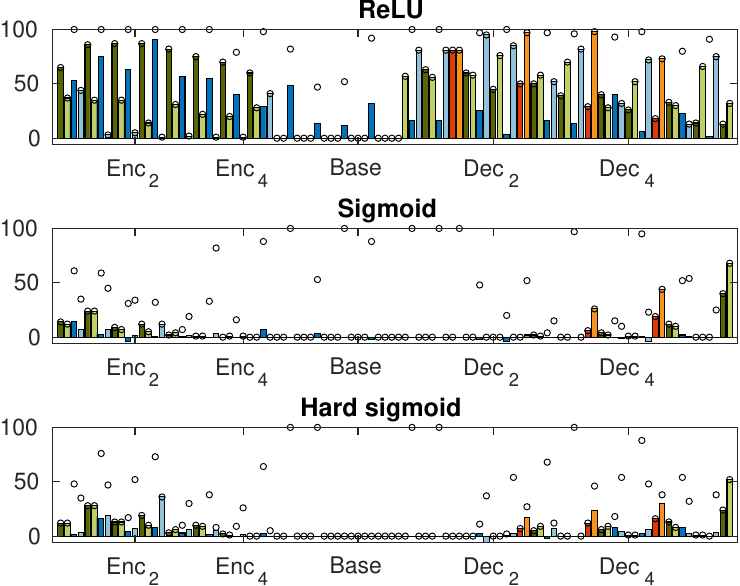}}
\caption{Difference between MSB bit-flip error rate and ratio of parameter set values in $1 < |x| < 2$. Unfilled dots indicate MSB bit-flip error rate. Pruned model.}
\label{fig:comparativaErrorRateMSBPruned}
\end{figure}

As shown in Fig. \ref{fig:comparativaErrorRateMSBPruned}, the central region of the ReLU-based DNN has the fewest errors, with most errors resulting from bit-flips in parameters within the range $1 < |x| < 2$.
This pattern also holds for bounded AFs, where almost all errors in the central region are due to this kind of bit-flips.

\subsection{Pruned quantized models}
Finally, the combined effects of applying pruning and full integer quantization are assessed.
Since this is a general analysis based on simulations, we used \textit{TensorFlow Lite}'s quantization scheme \cite{jacob2017quantization} to evaluate robustness against SBUs, while the AMD-Xilinx's Vitis AI tool quantization was applied for final implementation (see Section \ref{sec:quantization}).
The main difference is that \textit{TensorFlow Lite} uses a 32-bit integer representation instead of a 8-bit integer representation for biases (however, we verified that the variation in GIoU is below $\pm 0.08$ between both schemes).
According to \cite{jacob2017quantization}, the quantized version $\hat{r}$ of a real number $r$ is approximated by \eqref{eq:tfLiteQuantization}, where $S$ is a positive real scale factor, $q$ is an 8/32-bit integer value, and $Z$ is the zero-point, an integer value which is 0 for symmetric quantization.

\begin{equation}
    \hat{r} \approx r = S (q - Z)
    \label{eq:tfLiteQuantization}
\end{equation}

Thus, the value $q_y$ resulting from $Y = Wx + b$ is:

\begin{equation}
    q_y = \frac{S_w S_x}{S_y} (q_w q_x - Z_x q_w + q_b)
    \label{eq:tfLiteMAC}
\end{equation}

\begin{figure}[t]
\centerline{\includegraphics[height = 6.75cm]{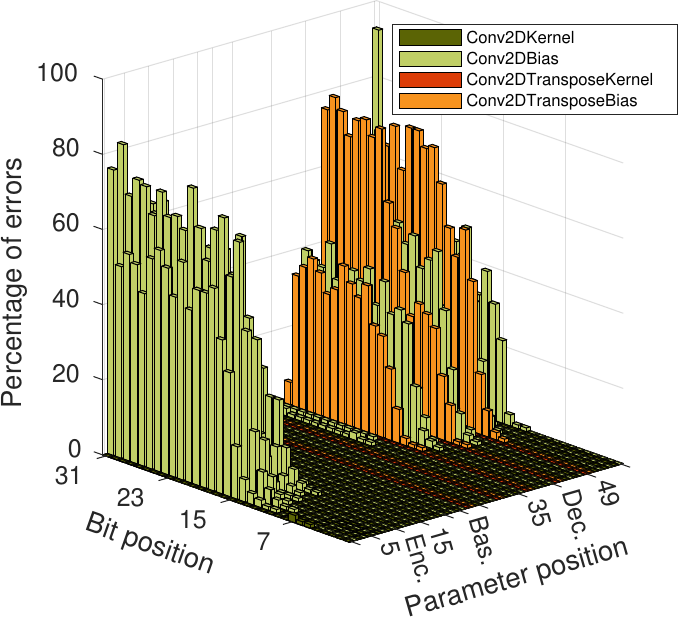}}
\caption{Bit-flip error rate for the pruned quantized ReLU-based DNN (weights are 8-bit long and biases are 32-bit long).}
\label{fig:bitFlipErrorPrunedQuantized}
\end{figure}

\begin{figure}[t]
\centerline{\includegraphics[height = 6cm]{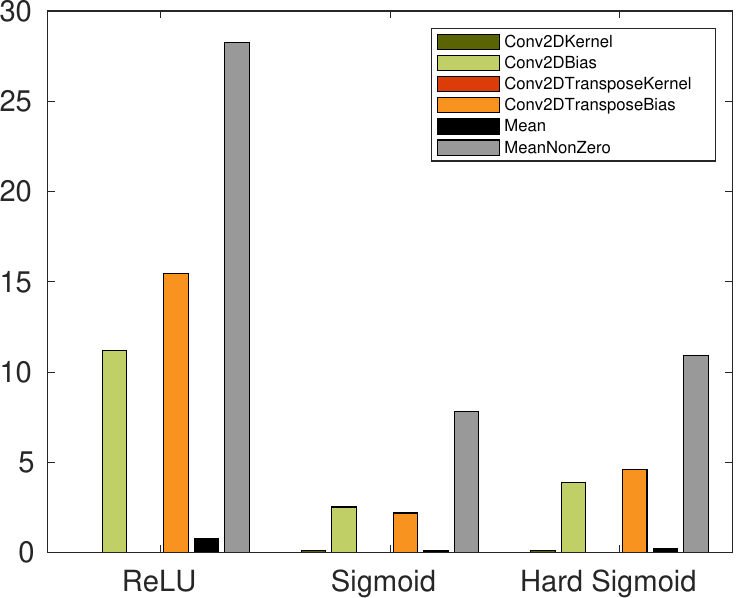}}
\caption{Mean bit-flip error rate in the pruned quantized model.}
\label{fig:comparativaErrorRateGeneralPrunedQuantized}
\end{figure}

Due to the per-tensor and mainly symmetric quantization scheme, the number of $q_i$ values to store is significantly greater than that of $Z_i$ and $S_i$ values, so bit-flips were only injected on $q_w$ and $q_b$.
The results for the pruned and quantized ReLU-based model are shown in Fig. \ref{fig:bitFlipErrorPrunedQuantized}.
Although the quantized model may initially seem less robust than its nonquantized counterpart, in this case only the biases are sensitive parameters.
Since biases make up an small part of the model, the overall proportion of sensitive parameters is much smaller than in the nonquantized model.
Regarding the comparison among AFs, bounded ones consistently offer greater robustness to the model as depicted in Fig. \ref{fig:comparativaErrorRateGeneralPrunedQuantized}.

\section{Deployment and performance characterization}\label{sec:characterization}
The DNNs were implemented on an AMD-Xilinx K26 SoM, which features an XCK26-SK-KV260-G Zynq UltraScale+ MPSoC containing a 64-bit Quad-Core ARM A-53 processor along with a 16nm FinFET Programmable Logic FPGA.
The DNNs have been deployed using AMD-Xilinx's Vitis AI 3.5 environment on a single-core B4096 Deep Processing Unit (DPU).
To characterize the performance of the models, the IoU on the test images, the throughput and the power consumption during DPU inference were measured on the KV260 SoM running Petalinux operating system (see Table \ref{tab:metricsPrunedQuantized}).
  
\begin{table}[t]
\caption{IoU of the pruned 8-bit integer DNNs on the test set.}
\label{tab:metricsPrunedQuantized}
\centering
\begin{tabular}{c|c|c|c|}
\hline
\multicolumn{1}{|c|}{\diagbox[]{\textbf{Class}}{\textbf{AF}}} & \multicolumn{1}{c|}{\textbf{ReLU}} & \multicolumn{1}{c|}{\textbf{Sigmoid}} & \multicolumn{1}{c|}{\textbf{Hard Sigmoid}} \\ \hline
\multicolumn{1}{|c|}{\textbf{Road}}       & \multicolumn{1}{c|}{97.72} & \multicolumn{1}{c|}{96.01} & \multicolumn{1}{c|}{96.74} \\ \hline
\multicolumn{1}{|c|}{\textbf{Marks}}      & \multicolumn{1}{c|}{88.15} & \multicolumn{1}{c|}{82.73} & \multicolumn{1}{c|}{81.07} \\ \hline
\multicolumn{1}{|c|}{\textbf{Vegetation}} & \multicolumn{1}{c|}{93.85} & \multicolumn{1}{c|}{88.94} & \multicolumn{1}{c|}{90.76} \\ \hline
\multicolumn{1}{|c|}{\textbf{Sky}}        & \multicolumn{1}{c|}{92.45} & \multicolumn{1}{c|}{82.82} & \multicolumn{1}{c|}{82.05} \\ \hline
\multicolumn{1}{|c|}{\textbf{Others}}     & \multicolumn{1}{c|}{77.38} & \multicolumn{1}{c|}{65.73} & \multicolumn{1}{c|}{69.95} \\ \hline
\multicolumn{1}{|c|}{\textbf{Global}}     & \multicolumn{1}{c|}{94.47} & \multicolumn{1}{c|}{90.67} & \multicolumn{1}{c|}{91.80} \\ \hline
\multicolumn{1}{|c|}{\textbf{Weighted}}   & \multicolumn{1}{c|}{88.37} & \multicolumn{1}{c|}{80.96} & \multicolumn{1}{c|}{80.72} \\ \hline
\end{tabular}
\end{table}

Throughput measurements were conducted by averaging the inference execution of 100 iterations on the test images.
The fact that the ReLU-based model was the most compressed after pruning, combined with the simplicity of calculating the AF, results in the highest throughput among the three, closely followed by the Hard Sigmoid-based model.
Since the Sigmoid function is not piecewise linear, its direct implementation on a DPU is unsupported and must be computed on a CPU core, thus dramatically augmenting the inference latency.

\begin{table}[b]
\caption{Characterization of the optimized models with different AFs on the KV260 SoM.}
\label{tab:modelsMetrics}
\centering
\begin{tabular}{|c|c|c|c|}
\hline
\textbf{\diagbox[]{Metric}{AF}} & \textbf{ReLU} & \textbf{Sigmoid}        & \textbf{Hard Sigmoid} \\ \hline
\textbf{Throughput (FPS)}       &   22.68       &  0.59$^{\mathrm{*}}$    &         19.83         \\ \hline
\textbf{Power (W)}              &   6.62        &  5.74$^{\mathrm{*}}$    &         6.90          \\ \hline
\textbf{Energy (J)}             &   0.29        &  9.73                   &         0.35          \\ \hline
\multicolumn{4}{l}{$^{\mathrm{*}}$Activation runs on CPU.}
\end{tabular}
\end{table}

Power consumption in the SoM was measured using the $tegrastats$ application thanks to the INA260 current sensor integrated in the KV260.
The Sigmoid-based DNN deployment consumes the least power since AFs are computed on the CPU, but its slow execution makes it the most energy-intensive, which is estimated by multiplying the inverse of the throughput by the mean power consumption.
On the contrary, the ReLU-based implementation is the most energy efficient.

\section{Conclusions}\label{sec:conclusions}
This article explores the use of bounded AFs in image segmentation DNNs to assess the robustness of these models against soft errors when deployed on embedded processing platforms for safety-critical applications.
As a general conclusion, we found that it is a trade-off between the resilience enhancements that bounded squashing AFs provide and the need for more aggressive model compression to improve computational performance and reduce memory footprint.

Regarding pruneability, ReLU-based and Hard Sigmoid-based models prove to have the larger pruneability factors, achieving a reduction in the number of parameters by over 95\% and a reduction in the number of GFLOPS by over 70\% in both cases.
When it comes to robustness against single bit-flips, bounded AFs show the best resilience as the propagation of generated perturbations throughout the model layers is notably reduced.
In 32-bit floating-point models, the most critical situation occurs when a bit-flip affects the MSB of the exponent for values in the range $1 < |x| < 2$, resulting in a conversion to $NaN$.
This primarily affects the Gamma parameters in Batch Normalization layers.
The over-parameterized nature of the original DNNs imply that as a model is pruned deeper, it seems to lose robustness.
However, the relative difference between ReLU-based and Hard Sigmoid-based models, both with similar pruning ratios, remains constant.
Nevertheless, the advisability of applying pruning techniques in relation to its influence over robustness must be contrasted with the particular design of the processor for deployment, since smaller models generally require fewer logic resources, decreasing the likelihood of an SBU occurring in a sensitive parameter.
For pruned and quantized integer models, the inherent binary representation prevents the occurrence of $NaNs$ or infinities, significantly reducing the error rate, while the relative robustness between different AFs show an identical pattern.

In terms of IoU, for DNNs implemented on an AMD-Xilinx KV260 SoM, the ReLU-based model achieved the best results.
This, combined with the computational complexity and high latencies of computing Sigmoid AF, makes the ReLU-based DNN the most efficient in terms of throughput and power consumption.
However, using Hard-Sigmoids as nodal AF deserves to be considered as a suitable design option for a good trade-off between robustness and performance.

\bibliography{biblio}
\bibliographystyle{ieeetr}
\end{document}